\documentclass[11pt,a4paper]{article}
\usepackage[hyperref]{acl2021}
\usepackage{times}
\usepackage{latexsym}

\usepackage{microtype}

\usepackage{url}
\usepackage{color,xcolor,colortbl}
\usepackage{graphicx}
\usepackage{amsmath}
\usepackage{multirow}
\usepackage{wrapfig}
\usepackage[utf8]{inputenc}

\definecolor{aliceblue}{rgb}{0.94, 0.97, 1.0}
\definecolor{beaublue}{rgb}{0.74, 0.83, 0.9}
\definecolor{lightblue}{rgb}{0.68, 0.85, 0.9}
\definecolor{lightgray}{rgb}{0.83, 0.83, 0.83}

\aclfinalcopy 
\def\aclpaperid{217} 

\title{More than just Frequency?\\Demasking Unsupervised Hypernymy Prediction Methods}

\author{Thomas Bott, Dominik Schlechtweg and Sabine Schulte im Walde \\
  Institute for Natural Language Processing, University of Stuttgart, Germany \\
  \texttt{\{bottts,schlecdk,schulte\}@ims.uni-stuttgart.de}\\}

\date{}

\begin{document}

\maketitle

\begin{abstract}
  This paper presents a comparison of unsupervised methods of
  hypernymy prediction (i.e., to predict which word in a pair of words
  such as \textit{fish--cod} is the hypernym and which the
  hyponym). Most importantly, we demonstrate across datasets for
  English and for German that the predictions of three methods (Weeds-
  Prec, invCL, SLQS Row) strongly overlap and are highly correlated
  with frequency-based predictions. In contrast, the second-order
  method SLQS shows an overall lower accuracy but makes correct
  predictions where the others go wrong. Our study once more confirms
  the general need to check the frequency bias of a computational
  method in order to identify frequency-(un)related effects.
\end{abstract}

\section{Introduction}

Hypernymy represents a major paradigmatic semantic relation between
two concepts, a hypernym (superordinate) and a hyponym (subordinate),
as in \textit{tree--oak} and \textit{fish--cod}, where the hyponym
implies the hypernym, but not vice versa. From a cognitive perspective
hypernymy is central to the organisation of the mental lexicon
\cite{Deese:65,Miller/Fellbaum:91,Murphy:03}, next to further semantic
relations such as synonymy, antonymy, etc. From a computational
perspective hypernymy is central to solving a number of Natural
Language Processing (NLP) tasks such as taxonomy creation
\cite{Hearst:98,CimianoEtAl:04,SnowEtAl:06,Navigli/Ponzetto:12},
textual entailment \cite{DaganEtAl:06,ClarkEtAl:07} and text generation \citep{Biran/McKeown:13}.

Accordingly, the field has witnessed active research on two subtasks
involved in computational models of hypernymy (see
\newcite{ShwartzEtAl:17} for an extensive overview): \textit{hypernymy
  detection} (i.e., distinguishing hypernymy from other semantic
relations) and \textit{hypernymy prediction} (i.e., determining which
word in a pair of words is the hypernym and which is the hyponym). The
target subtask of the current study is hypernymy prediction: we
perform a comparative analysis of a class of approaches commonly
refered to as \textit{unsupervised hypernymy methods}
\cite{WeedsEtAl:04,KotlermanEtAl:10,Clarke:11,Lenci/Benotto:12,SantusEtAl:14}. These
methods all rely on the distributional hypothesis
\cite{Harris:54,Firth:57} that words which are similar in meaning also
occur in similar linguistic distributions. In this vein, they exploit
asymmetries in distributional vector space representations, in order
to contrast hypernym and hyponym vectors.

While these unsupervised hypernymy prediction methods have been
explored and compared extensively on a number of benchmark datasets
\citep{ShwartzEtAl:17}, this study takes a novel perspective and
performs a detailed analysis of whether and where the methods make
similar or different decisions. Our prediction experiments on simplex
and complex nouns in English and German WordNets and evaluation
benchmarks show that most of the methods we investigate overlap in
their specific predictions to a surprisingly high degree, and that the
predictions strongly correlate with those based on raw
frequencies. Our study therefore emphasises the general need to check
the frequency bias of a computational method and to distinguish
between frequency-related and frequency-unrelated effects.

\section{Data and Methods}

In the following we describe our gold standard datasets (Section~\ref{sec:gs}), our corpora and vector spaces (Section~\ref{sec:spaces}) and our hypernymy prediction methods (Section~\ref{sec:methods}). The code and links to the gold standards are available from
\url{https://github.com/Thommy96/hyp-freq-comp}.

\subsection{Gold Standard Datasets}
\label{sec:gs}

Our study focuses on hypernymy between nouns and uses two types of
gold standard resources for hypernymy relations. On the one hand, we
rely on WordNets as classical large-scale taxonomies where hypernymy
represents one of the core semantic relations for organisation: the
English \textit{WordNet}\footnote{\url{https://wordnet.princeton.edu}}
\cite{MillerEtAl:90,Fellbaum:98}, version~3, and the German
\textit{GermaNet}\footnote{\url{https://uni-tuebingen.de/en/142806}}
\cite{Hamp/Feldweg:97,Kunze/Wagner:99,Lemnitzer/Kunze:07},
version~11. From both WordNets, we extracted all noun--noun pairs with
a hypernymy relation and removed duplicates, autohyponyms and
space-separated multiword expressions.  We also distinguish between
compounds (which frequently represent hyponyms of their constituent
heads, as in \textit{dog--lapdog}) and non-compounds by applying a
simple heuristic, i.e., categorising all hypernym--hyponym pairs as
compounds if one is a substring of the other.
We expected this subset to exhibit idiosynchratic behaviour in our
prediction experiments.

On the other hand, we rely on a number of benchmark datasets for
hypernymy evaluation: \textit{BLESS} \cite{Baroni/Lenci:11} provides
related concepts for 200 English concrete nouns connected through a
semantic relation (hypernymy, co-hyponymy, meronymy, attribute, event)
or a null-relation. The dataset by \newcite{Lenci/Benotto:12} contains
a subset of BLESS relation pairs, as created for previous comparisons
of hypernymy detection methods. A dataset similar to BLESS,
\textit{EVALution}, was induced from ConceptNet and WordNet
\cite{SantusEtAl:15}. Its semantic relations include hypernymy,
synonymy, antonymy and meronymy. For quality reasons, the pairs were
filtered by automatic methods and crowd-sourcing to improve
consistency and to determine prototypical pairs. Finally, we use the
Weeds dataset \cite{WeedsEtAl:04,Weeds/Weir:05} which contains word
pairs related by hypernymy and co-hyponymy across word classes. From
all four benchmark datasets we extracted all noun--noun pairs related
by hypernymy.

The first row in Table~\ref{tab:results} shows the numbers of
hypernymy pairs in the WordNets and in the benchmark datasets.

\subsection{Corpora and Vector Spaces}
\label{sec:spaces}

We created our distributional vector spaces based on the
\textit{WaCky}\footnote{\url{http://wacky.sslmit.unibo.it/}} corpora
\cite{BaroniEtAl:09} for English and for German. The English
\textit{PukWaC} corpus is the syntax-annotated version of
\textit{ukWaC} \cite{FerraresiEtAl:08} and contains $\approx$1.9
billion words; the German \textit{SdeWaC} corpus
\cite{Faass/Eckart:13} is a cleaned version of the \textit{WaCky}
corpus \textit{deWaC} and contains $\approx$880 million words; both
corpora are pos-tagged with the \textit{TreeTagger} \cite{Schmid:94}.

For each corpus we created a traditional count vector
space\footnote{Note that not all of the selected methods are
  applicable to embeddings, and it also not our goal to identify the
  optimal vector spaces, rather than analysing their predictions; this
  is why our analyses rely on standard count dimensions.} based on a
co-occurrence window of $\pm$~10 words within sentences (because
sentences in the SdeWaC are shuffled, so going beyond sentence border
is meaningless). We used a bag-of-words approach only taking into
account lemmatised nouns, verbs and adjectives.

\begin{table*}[t!]
  \centering
  \begin{tabular}{|l||c|c||c|c||c|c|c|c|} \hline

    & \multicolumn{2}{c||}{\textbf{WordNet}} & \multicolumn{2}{c||}{\textbf{GermaNet}} & \multirow{2}*{\textbf{BLESS}} & \multirow{2}*{\textbf{EVALution}} & \multirow{2}*{\textbf{LB}} & \multirow{2}*{\textbf{Weeds}} \\ \cline{2-5}

    & $\neg$comp & comp & $\neg$comp & comp & & & & \\ \hline \hline

    \multicolumn{1}{|r||}{\textit{sizes:}} & 106,397 & 3,366 & 102,714 & 35,963 & 1,337 & 606 & 1,747 & 1,117 \\ \hline \hline

    \textbf{Word Length} & \cellcolor{aliceblue}47.26 & \cellcolor{lightgray}\textbf{94.65} & \cellcolor{aliceblue}56.14 & \cellcolor{lightgray}\textbf{99.41} & \cellcolor{aliceblue}23.19 & \cellcolor{aliceblue}34.86 & \cellcolor{aliceblue}52.42 & \cellcolor{aliceblue}44.76 \\ \hline
    \textbf{Word Frequency} & \cellcolor{aliceblue}\textbf{73.19} & \cellcolor{lightgray}92.81 & \cellcolor{aliceblue}73.66 & \cellcolor{lightgray}98.78 & \cellcolor{aliceblue}62.30 & \cellcolor{aliceblue}\textbf{68.96} & \cellcolor{aliceblue}76.48 & \cellcolor{aliceblue}\textbf{76.63} \\ \hline \hline

    \textbf{WeedsPrec} & \cellcolor{aliceblue}72.22 & \cellcolor{lightgray}92.93 & \cellcolor{aliceblue}74.01 & \cellcolor{lightgray}98.87 & \cellcolor{aliceblue}57.52 & \cellcolor{aliceblue}64.22 & \cellcolor{aliceblue}\textbf{77.02} & \cellcolor{aliceblue}74.22 \\ \hline
    \textbf{InvCL} & \cellcolor{aliceblue}72.97 & \cellcolor{lightgray}92.84 & \cellcolor{aliceblue}73.92 & \cellcolor{lightgray}98.78 & \cellcolor{aliceblue}63.05 & \cellcolor{aliceblue}68.86 & \cellcolor{aliceblue}76.48 & \cellcolor{aliceblue}76.45 \\ \hline
    \textbf{SLQS Row} & \cellcolor{aliceblue}71.82 & \cellcolor{lightgray}93.02 & \cellcolor{aliceblue}\textbf{74.40} & \cellcolor{lightgray}98.79 & \cellcolor{aliceblue}58.56 & \cellcolor{aliceblue}55.91 & \cellcolor{aliceblue}71.27 & \cellcolor{aliceblue}72.43 \\ \hline
    \textbf{SLQS Sec} & \cellcolor{aliceblue}65.05 & \cellcolor{lightgray}74.66 & \cellcolor{aliceblue}70.38 & \cellcolor{lightgray}90.15 & \cellcolor{aliceblue}\textbf{71.80} & \cellcolor{aliceblue}59.63 & \cellcolor{aliceblue}62.66 & \cellcolor{aliceblue}65.71 \\ \hline

  \end{tabular}
  \caption{Sizes of datasets and overall prediction results across datasets.}
  \label{tab:results} 
\end{table*}

\vspace{+2mm}
\subsection{Hypernymy Methods and Baselines}
\label{sec:methods}

We selected four unsupervised hypernymy methods and defined two
baselines. The methods were chosen from different families with regard
to how they exploit the distributional hypothesis for hypernymy
detection: \textit{WeedsPrec} and \textit{InvCL} rely on the
\textit{Distributional Inclusion Hypothesis}, according to which a
significant number of distributional features of a word $x$ is
included in the distributional features of a word $y$, if $x$ is
semantically more specific than~$y$. \textit{SLQS Row} and
\textit{SLQS~Sec}\footnote{Originally, this method is called SLQS, but
  to distinguish it from SLQS Row we refer to it as SLQS~Sec.}  rely
on the \textit{Distributional Informativeness Hypothesis} using first-
and second-order variants of word entropy, respectively. The methods
are defined as follows regarding the distributional features $f$ in
the two word vectors $\vec{x}$ and $\vec{y}$ for a word pair
$\langle x,y \rangle$.
\vspace{+4mm}\\
\textit{\textbf{WeedsPrec}}: An asymmetric precision method suggested
by \newcite{WeedsEtAl:04} that quantifies the weighted inclusion of
the features of word $x$ in the features of word $y$. If
$WeedsPrec(x,y) > WeedsPrec(y,x)$, then $x$ is predicted as the
hyponym and $y$ as the hypernym, and vice versa.
\begin{align*}
  WeedsPrec(x,y) = \frac{\sum_{f \in (\overrightarrow{x} \cap \overrightarrow{y})} \; x_f}{\sum_{f \in \overrightarrow{x}} \; x_f}
\end{align*}
\textit{\textbf{InvCL}}: An asymmetric precision method suggested by
\newcite{Lenci/Benotto:12} that takes both feature inclusion as well
as feature non-inclusion (originally suggested as \textit{ClarkDE
  (cde)} by \newcite{Clarke:11}) into account. If
$invCL(x,y) > invCL(y,x)$, then $x$ is predicted as the hyponym and
$y$ as the hypernym, and vice versa.

\vspace{-1mm}
\begin{align*}
  cde(x,y) = \frac{\sum_{f \in (\overrightarrow{x} \cap \overrightarrow{y})} min(x_f,y_f)}{\sum_{f \in \overrightarrow{x}} \; x_f} 
\end{align*}
\vspace{-2mm}
\begin{align*}
  invCL(x,y) = \sqrt{cde(x,y) \cdot (1-cde(y,x))}
\end{align*}
\vspace{+1mm}\\
\textit{\textbf{SLQS Row}}: An asymmetric method suggested by
\newcite{ShwartzEtAl:16} which relies on the word entropy $H(w)$ for a
word $w$, taking all context words as features into account: $w_f$. If
$SLQS_{Row}(x,y)~>~0$, then $x$ is predicted as the hyponym and $y$ as
the hypernym, and vice versa.
\begin{align*}    
  SLQS_{Row}(x,y) = 1-\frac{H(x)}{H(y)}\\
  H(w) = -\sum_{w_f} \; p(w_f|w) \cdot log_{2}(p(w_f|w))
\end{align*}
\vspace{+1mm}\\
\textit{\textbf{SLQS Sec}}: An asymmetric method suggested by
\newcite{SantusEtAl:14} which relies on second-order word entropy
$E(w)$ and is calculated as the median entropy $Med$ of a word's most
strongly associated context words $w_f$. We use the 50 strongest
contexts in our vector spaces, as determined by weighted co-occurrence
scores using \textit{positive local mutual information}
\cite{Evert:05}. If $SLQS_{Sec}(x,y) > 0$, then $x$ is predicted as
the hyponym and $y$ as the hypernym, and vice versa.
\begin{align*}
  SLQS_{Sec}(x,y) = 1-\frac{E(x)}{E(y)}\\
  E(w) = Med_{w_f} \; H(w_f)
\end{align*}
%
\textbf{Baselines:} In comparison to the hypernymy methods we applied
two baselines, cf. Zipf's principles of least effort \citep{zipf49}:
\begin{itemize}
\item \textit{\textbf{Word Length}}: Given that hyponyms refer to more
specific concepts than their hypernyms, and assuming that more
specific concepts tend to have a longer word length, this baseline
predicts the longer word in a word pair (as measured by the number of
characters) as the hyponym.
\item \textit{\textbf{Word Frequency}}: Given that hyponyms refer to more
specific concepts than their hypernyms and assuming that more specific
concepts appear less often in a corpus, this baseline predicts the
less frequent word in a word pair (as measured by corpus frequency) as
the hyponym.
\end{itemize}

\begin{figure*}[h!]
  \centering
  \includegraphics[width=0.65\linewidth]{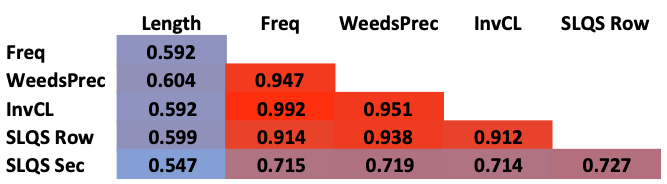}
  \vspace{+5mm}\\
  \includegraphics[width=0.65\linewidth]{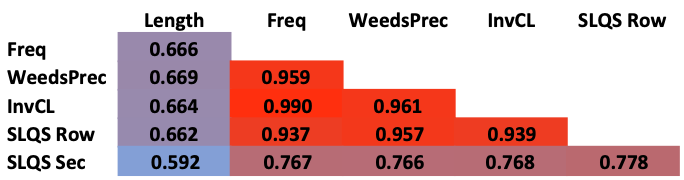}
  \caption{SMC correlations between methods for WordNet (above) and
    GermaNet (below) non-compound pairs.}
  \label{fig:smc}
\end{figure*}

\section{Prediction Results and Comparisons}

\subsection{Overall results}

Table~\ref{tab:results} shows the overall accuracy results of the
predictions across methods and datasets (best results in bold
fonts). Accuracy is defined by the proportion of correct predictions
given that we know which word in a word pair is the hypernym and which
is the hyponym.

For each WordNet we list two results, one for the non-compound
pairs (in blue, as the benchmark results) and one for the compound
pairs (in grey). For compound pairs word length is an almost perfect
predictor,\footnote{The prediction does not reach 100\% because our
  heuristic included non-compound pairs, such as
  \textit{selection--election}.} as expected, and all unsupervised
methods are also above 90\%, with SLQS~Sec as an exception.
In all other columns we can see that word length is generally a poor
baseline. Word frequency, however, is a very powerful baseline; across
datasets it keeps up or even outperforms the respective best methods,
which are SLQS~Sec on BLESS; InvCL on EVALution and Weeds; and
WeedsPrec on LB. Across
datasets, the best results vary between 68.96\% and 77.02\% for
non-compounds; compounds obviously represent ``easy'' cases of
hypernymy.

\subsection{Correlations between predictions}
\label{sec:smc}

To explore similarities in predictions across methods, we applied the
\textit{Simple Matching Coefficient (SMC)} \cite{Sokal:58} to
determine for each two methods to which degree their decisions
overlap, by comparing the number of matching decisions (i.e., where
both methods predicted the same noun in a noun pair as the hypernym)
against the number of decisions (i.e., the total number of noun
pairs). The heatmaps in Figure~\ref{fig:smc} show the results for the
non-compounds in the English WordNet (left) and in GermaNet
(right). They clearly demonstrate that word length makes very
different decisions to word frequency and the unsupervised methods,
and that word frequency and all unsupervised methods but SLQS~Sec
highly correlate in their predictions.

\begin{figure*}[ht!]
  \centering
  \includegraphics[width=0.65\linewidth]{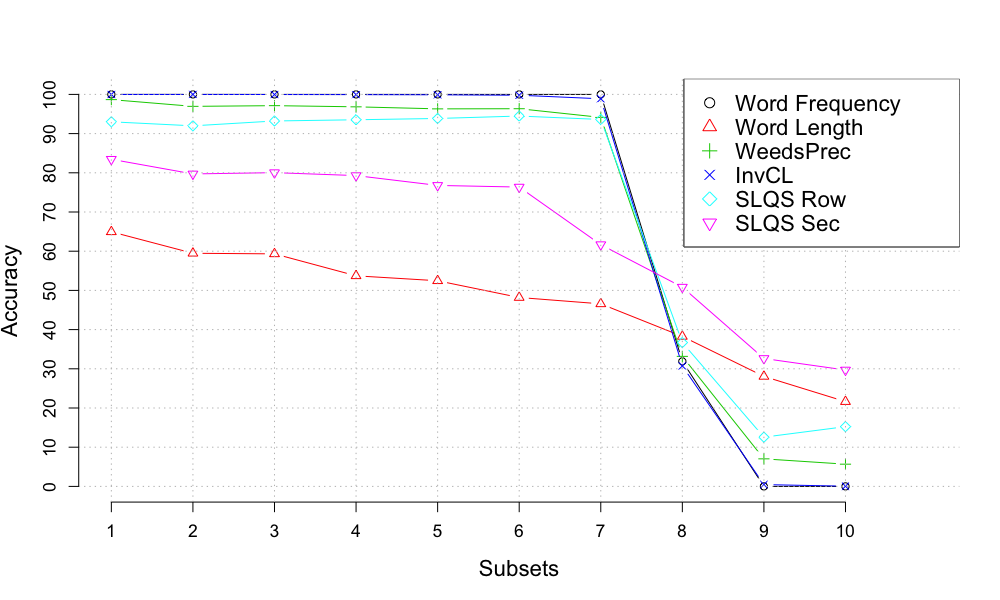}
  \vspace{+1mm}\\
  \includegraphics[width=0.65\linewidth]{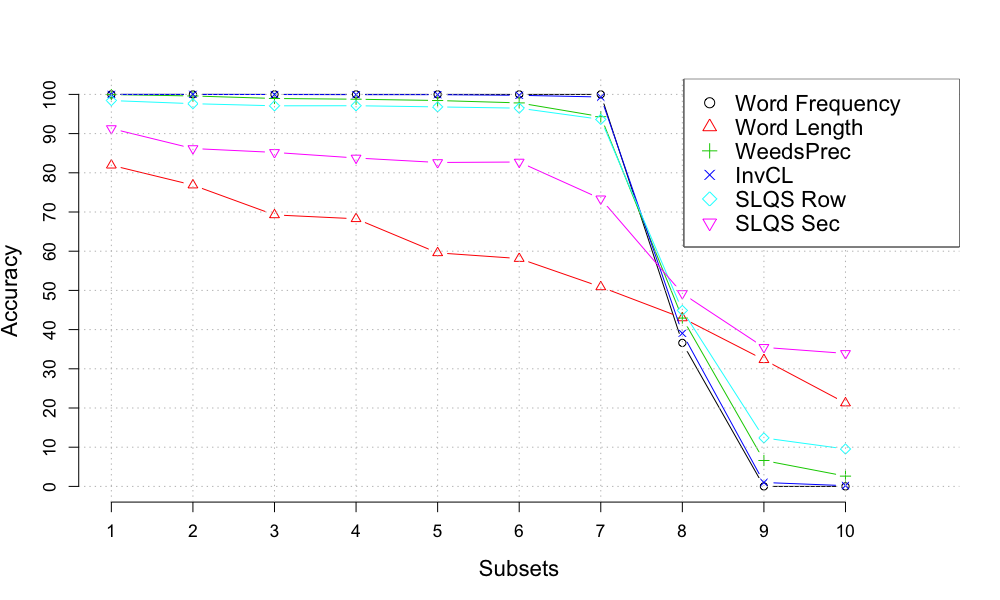}
  \caption{Prediction results for WordNet (above) and GermaNet (below)
    non-compound pairs across equally-sized subsets of target pairs
    sorted by difference in hypernym frequency and hyponym frequency.}
  \label{fig:results-freq-diff}
\end{figure*}

\subsection{Role of frequency}

We go one step further to explore the role of
frequency. Figure~\ref{fig:results-freq-diff} presents the prediction
results on 10 equally-sized subsets of the non-compound pairs in the
WordNets after the target pairs were sorted by decreasing difference
in hypernym corpus frequency minus hyponym corpus frequency. I.e., in
the left-most subset on the $x$-axis we see the results on the subset
with largest differences in hypernym--hyponym frequencies.

We can clearly see that up to subset 7 (up to which the hyponym
frequencies are all below the hypernym frequencies), decisions based
on word frequency, WeedsPrec, invCL and SLQS~Row predict the hypernym
almost perfectly; for subset 8 (where the hyponym frequencies start to
become larger than the hypernym frequencies) their predictions are
becoming worse; and for subsets 9--10 the predictions are mostly
wrong. Results by relying on word length and SLQS~Sec are clearly
worse for the first seven subsets but also better for the last two
subsets, thus confirming that they make different predictions.

\subsection{Correctness of predictions}

\begin{figure}[h!]
  \centering
  \includegraphics[width=1.05\linewidth]{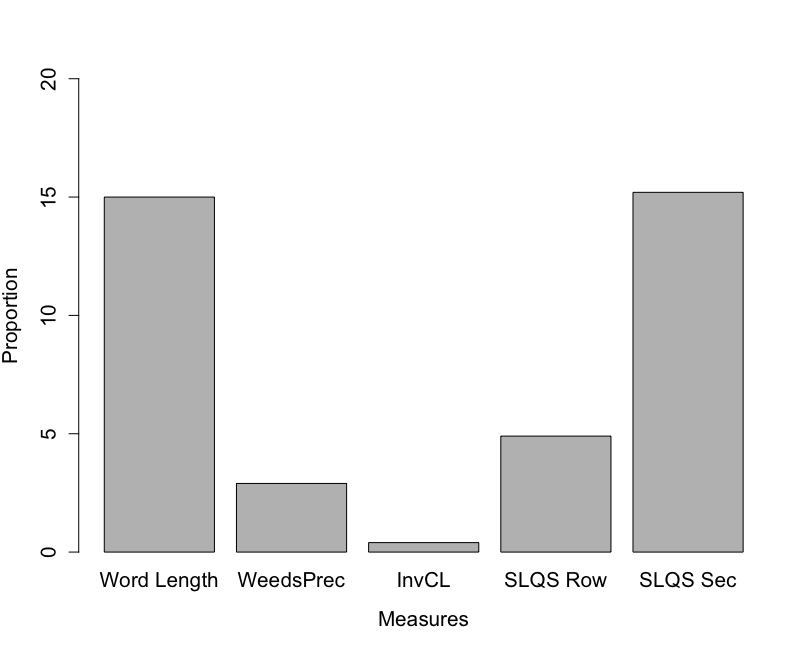}
  \vspace{-2mm}
  \caption{Proportions of pairs predicted wrongly by word frequency but correctly by the given method.}
  \label{fig:prop-right-wrong}
\end{figure}

While SMC in Section~\ref{sec:smc} informed us about overlap in
decisions, it did not tell us whether one of the methods is
qualitatively superior, so we analysed whether some methods are simply
worse than others, according to their lower accuracy in prediction, or
whether the methods all have their own strengths. We calculated for
each pair of methods which proportion of wrongly predicted pairs of
one method was predicted correctly by the other
method. Figure~\ref{fig:prop-right-wrong} illustrates for the English
WordNet how many pairs wrongly predicted by word frequency are
predicted correctly by another method (see $x$-axis).

As we can see, while word length and SLQS~Sec are often worse in
performance than frequency, they still manage to make correct
predictions when frequency fails, which is much less the case for the
frequency-alike methods WeedsPrec, invCL and SLQS~Row. In particular,
invCL seems to make almost identical predictions as frequency, which
was already indicated by their almost perfectly overlapping lines in
Figure \ref{fig:results-freq-diff}.

\section{Conclusion}

This study performed a series of hypernymy predictions by unsupervised
methods. We demonstrated that across datasets for English and for
German the predictions of three methods (WeedsPrec, inv- CL and
SLQS~Row) are highly correlated and also mostly identical with
frequency-based predictions. In contrast, word length and SLQS~Sec
show an overall lower accuracy but at the same time make correct
predictions where the others go wrong. Our study once more confirms
the general need to check the frequency bias of a computational method
in order to identify frequency-(un)related effects.

\section*{Acknowledgements}

Dominik Schlechtweg was supported by the Konrad Adenauer Foundation
and the CRETA center funded by the German Ministry for Education and
Research during the conduct of this research.


\bibliographystyle{acl_natbib}
\bibliography{ssiwBib}

\end{document}